\documentclass[journal]{IEEEtran}
\usepackage{url}
\usepackage{tikz}
\usetikzlibrary{shapes.geometric}
\ifCLASSINFOpdf
  \usepackage{graphicx}
\usepackage{tikz}
\tikzset{ell/.style={draw, shape=ellipse}}
\else
\fi
\begin{document}
%
\title{The Era of End-to-End Autonomy:
Transitioning from Rule-Based Driving to Large Driving Models
}
%
%
%

\author{Prof. Em. Eduardo Nebot,~\IEEEmembership{Fellow,~IEEE,}
        and J. Stephany Berrio Perez ,~\IEEEmembership{Member,~IEEE}
\thanks{The Australian Centre for Robotics, School of Aeronautical, Mechanical and Mechatronic Engineering, The University of Sydney, Australia}}

%
%

\markboth{Preprint, March~2026}%
{Shell \MakeLowercase{\textit{et al.}}: Bare Demo of IEEEtran.cls for IEEE Journals}
%



\maketitle

\begin{abstract}

Autonomous driving is undergoing a shift from modular rule-based pipelines toward end-to-end (E2E) learning systems. This paper examines this transition by tracing the evolution from classical sense–perceive–plan–control architectures to large driving models (LDMs) capable of mapping raw sensor input directly to driving actions. We analyze recent developments including Tesla’s Full Self-Driving (FSD) V12–V14, Rivian’s Unified Intelligence platform, NVIDIA Cosmos, and emerging commercial robotaxi deployments, focusing on architectural design, deployment strategies, safety considerations and industry implications. A key emerging product category is supervised E2E driving, often referred to as FSD (Supervised) or L2++, which several manufacturers plan to deploy from 2026 onwards. These systems can perform most of the Dynamic Driving Task (DDT) in complex environments while requiring human supervision, shifting the driver’s role to safety oversight. Early operational evidence suggests E2E learning handles the long-tail distribution of real-world driving scenarios and is becoming a dominant commercial strategy. We also discuss how similar architectural advances may extend beyond autonomous vehicles (AV) to other embodied AI systems, including humanoid robotics.

\end{abstract}


%
\IEEEpeerreviewmaketitle

\section{Introduction}
%
%
%
%
\IEEEPARstart{T}{he} development of autonomous vehicles (AVs) has been one of the defining technological challenges of the twenty-first century. For more than two decades, the dominant paradigm was a rule-based modular pipeline in which specialized subsystems handled sensing, perception, prediction, planning, and control independently \cite{liu2025autonomous}. Each module was individually engineered, extensively hand-tuned, and interconnected through carefully specified interfaces. Although this approach delivered impressive results in structured environments \cite{DurrantWhyte2007StraddleCarrier} \cite{FutureBridgeAutonomousMining}, it became increasingly brittle when confronted with the combinatorial complexity of real-world urban traffic, the so-called long-tail problem \cite{Liu2024CurseOfRarity}.
By 2024-2025, a decisive architectural transition had taken shape. End-to-end (E2E) neural networks, trained on vast fleets of real-world driving data, had begun to outperform modular stacks in terms of performance and passenger comfort metrics. Tesla’s Full Self-Driving (FSD) systems, having transitioned to true E2E operation with version 12, further expanded the paradigm with versions 13 and 14, incorporating audio-based environmental awareness, multi-second temporal reasoning, and potentially a mixture-of-models in its latest FSD software version \cite{Elluswamy2025ICCV}. In early 2026, NVIDIA unveiled an ecosystem for designing, simulating, and validating autonomous driving in urban scenarios. The platform includes a world foundation model, a large-scale dataset for evaluation, and an open-source autonomous agent \cite{Huang2026NvidiaCES}. More recently, Rivian \cite{ScaringeRivianAutonomy} presented its roadmap to autonomy, emphasizing end-to-end (E2E) approaches to training Large Driving Models using data collected from its fleet. In June 2025, Waymo and Tesla launched commercial robotaxi services in Austin, Texas. Waymo has been operating a sensor-rich approach over several years, whereas Tesla, for the first time, deployed production vehicles using a camera-only (E2E) strategy. Tesla successful demonstrations helped pave the way for a larger deployment of a new capability often referred to as FSD (Supervised), commonly described as “L2++” \cite{reuters2025tesla_waymo_robotaxi}. In practical terms, the system can perform most of the Dynamic Driving Task (DDT) in urban environments, shifting the human role primarily to supervision. Furthermore, automotive OEMs have become increasingly proactive and have begun forming joint ventures with providers of E2E driving technology, such as Rivian–Volkswagen \cite{RivianVW2025} and Mercedes-Benz–NVIDIA \cite{Huang2026NvidiaCES} with plans to start to deploy L2++ for passenger cars in 2026 and potentially L4 in the near future.

This paper provides an analysis of the transition to E2E, including a discussion of the technology implications of the wide acceptance of supervised FSD for drivers and OEMs . Section 2 reviews the traditional autonomous driving architecture and its limitations. Section 3 introduces the E2E learning paradigm and large driving models. Section 4 analyzes the robotaxi landscape, comparing Waymo and Tesla’s approaches focusing on the recent deployment in Austin, Texas, in June 2025.  Section 5 presents a full analysis of FSD supervised widely deployed by Tesla across the latest E2E FSD version V13-V14. It also included a discussion of similar approaches from other OEMs. Section 6 explores the extension of E2E intelligence to humanoid robotics and simulation platforms, and Section 7 offers conclusions.

\section{Traditional Autonomous Driving Architecture}

\subsection{The Modular Pipeline}

\begin{figure}
    \centering
    \includegraphics[trim={6cm 0 0 0},clip, width=\linewidth]{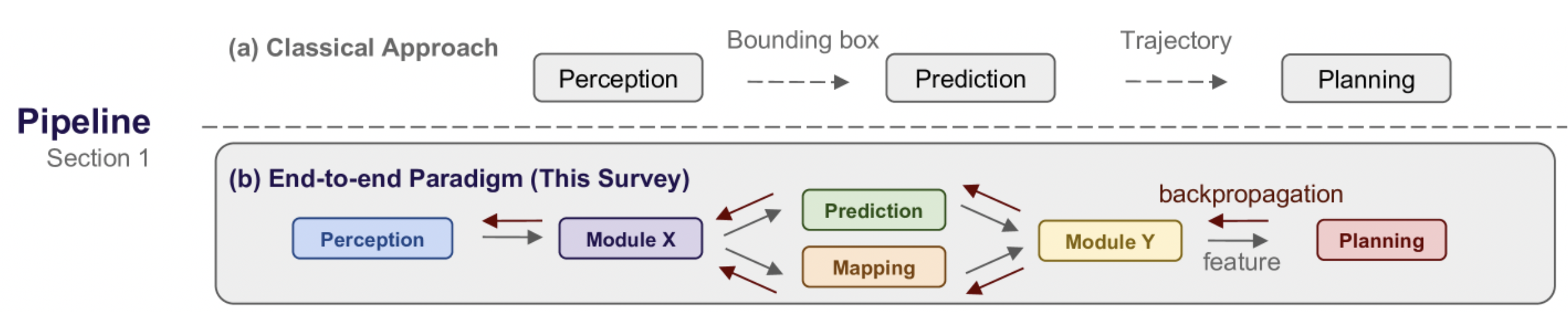}
    \caption{Authors in  \cite{10.1109/TPAMI.2024.3435937} define (a) the classical modular approach separates perception, prediction, and planning through intermediate representations such as bounding boxes and trajectories. (b) The end-to-end paradigm jointly learns interconnected modules, allowing information flow and backpropagation across perception, mapping, prediction, and planning components.}
    \label{fig:end_to_end_paper}
    \vspace{-4mm}
\end{figure}

The classical autonomous driving stack is typically organized as a sequential pipeline of functional modules (see Figure \ref{fig:end_to_end_paper}a). The sensing layer collects raw data from onboard sensors such as cameras, LiDAR, radar, ultrasonic sensors, wheel encoders, inertial measurement units (IMUs), and GNSS receivers, providing complementary observations of the vehicle and its surroundings \cite{ibanezguzman2012autonomous}.
A mapping and localization layer integrates these measurements with prior map information to maintain a consistent spatial representation of the environment and estimate the precise pose of the ego vehicle relative to a local or high-definition (HD) map \cite{Wijaya2024HighDM}.
The perception module processes sensor data to detect and classify objects, identify drivable areas, and estimate the positions and velocities of surrounding agents. The prediction module then forecasts the future trajectories of these agents on a horizon of several seconds \cite{liang2020pnpnet}.
Using this information, the planning module generates a safe and comfortable trajectory for the ego vehicle, while the control layer converts the planned trajectory into actuation commands for the steering, throttle, and braking systems.
In practice, each module may combine signal processing, computer vision, and machine learning methods. This modular architecture simplifies debugging, enables targeted improvements, and allows components to be replaced independently. HD maps often play a central role by providing detailed geometric and semantic information that reduces the perception burden on onboard algorithms and improves reliability.

\subsection{Limitations of Rule-Based Approaches}
Despite its conceptual clarity, the modular rule-based architecture has several fundamental limitations \cite{Leong2022ModularEndToEndAV}. First, the long-tail problem: the space of possible driving scenarios is effectively unbounded, making it impractical to hand-code rules for every edge case. Engineers may spend years adding heuristics for rare situations that human drivers handle intuitively and conservatively. Second, sensor and infrastructure dependence: many systems rely on expensive sensor suite, particularly multi-beam LiDAR, radar, and continuously updated HD maps, which increases system cost and complexity \cite{technologies11050117}. Third, maintenance overhead: ensuring proper calibration \cite{Meruva2020SensorCalibrationAV} and operation of such complex systems requires significant ongoing effort. While these costs may be manageable for commercial fleets, they make large-scale deployment in consumer vehicles far less practical.

\section{The Shift to End-to-End Learning and Large Driving Models}

\subsection{The End-to-End Paradigm}

End-to-end learning approaches the driving task as a single optimization problem. Rather than decomposing the task into specialized modules, an E2E model receives raw sensor inputs, typically camera images, kinematic information, maps, desired trajectory and in some architectures also radar, laser, audio  and directly outputs steering angles and acceleration commands. The entire model is trained jointly, allowing the gradients to flow from the output loss all the way back to the raw input representations (see Figure \ref{fig:end_to_end_paper}b). In principle, this allows the model to find intermediate representations that are optimally suited to the full driving task, rather than to any particular subtask.
Early E2E approaches, such as NVIDIA’s DAVE \cite{NetScale2004OffRoad} system in 2004 and its successor \cite{Bojarski2016EndToEnd} in 2016, demonstrated that a convolutional neural network could learn to steer a vehicle from camera input alone. However, these systems were limited to relatively simple highway scenarios. The key insight driving the renaissance of E2E learning in the 2020s is scale: when trained on millions of hours of real-world driving data collected from large vehicle fleets, E2E models exhibit emergent behaviors that were never explicitly programmed, including appropriate responses to unusual pedestrian behavior, construction zones, emergency vehicles, and adverse weather conditions.

\subsection{Large Driving Models}
The concept of Large Driving Models (LDMs) draws an explicit analogy with Large Language Models (LLMs) in natural language processing. Just as GPT-style transformers acquire broad linguistic competence from internet-scale text and other types of data, LDMs acquire broad driving competence from fleet-scale driving data. Key architectural features include transformer-based sequence modeling for temporal reasoning over multiple seconds of driving history \cite{shao2024lmdrive}, tokenized/diffused representations of scenes that enable attention across spatial and temporal dimensions, and the use of imitation learning and reinforcement learning to align model behavior with human driver preferences and safety constraints.
LDMs offer several advantages over both modular systems and earlier small-scale E2E models. They generalize more robustly to novel scenarios because they have been exposed to a vastly larger distribution of real-world situations during training. They can leverage weak supervision signals at scale, such as intervention events where a human driver corrects the system, without requiring expensive manual labeling of every training frame. The model can be continuously fine-tuned as new data is collected from the deployed fleet, allowing a cycle of improvement through the updates of the different models version. This is currently done progressively, using selected data that are automatically detected as relevant for additional training. The system is then retrained using reinforcement learning to learn safer behaviors, and the updated policy is deployed to the fleet after the model is distilled.

\subsubsection{Two-step training: from “good human driver” to safer-than-human robustness}

\begin{figure}[ht]
    \centering

\resizebox{0.9\columnwidth}{!}{%

\begin{tikzpicture}[
  font=\sffamily,
  every node/.style={align=center, text=white},
  box/.style={draw=black, very thick, rounded corners=18pt, inner sep=1pt},
  ell/.style={draw=black, very thick, ellipse, inner sep=1pt},
  x=0.7cm, y=1cm
]

\definecolor{darkgreen}{HTML}{1B6B2A}
\definecolor{darkblue}{HTML}{0F5F7A}

\node[box, fill=darkgreen, minimum width=7cm, minimum height=2.2cm] (il) at (0,2.2)
{Imitation Learning\\[3pt]Curated Dataset};

\node[box, fill=darkblue, minimum width=7cm, minimum height=2.4cm] (rl) at (0,-0.2)
{Reinforcing Learning\\[6pt]
New data from Fleet: Edge cases, \\
human, values, \ldots.};

\node[ell, fill=darkgreen, minimum width=4.8cm, minimum height=1.7cm] (base) at (10.2,2.2)
{Baseline\\Driving Policy};

\node[ell, fill=darkblue, minimum width=5.5cm, minimum height=2.3cm] (new) at (10.2,-0.2)
{New Safer Behaviors\\Model Updates};

\end{tikzpicture}
}
\caption{Training Process for large Driving Models. Phase one use a curated dataset based on good driving behaviors to obtain a baseline model. In Phase 2, additional data, comprising fleet-collected edge cases and synthetically generated scenarios, are used to further train the policy via reinforcement learning, yielding an updated model with improved safety.}
\label{fig:driving_model}
\vspace{-4mm}

\end{figure}
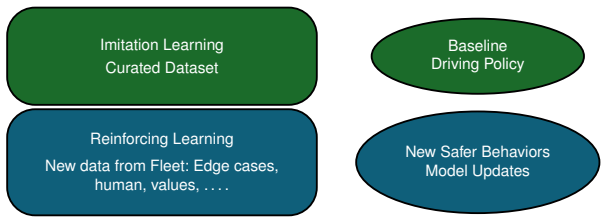

End-to-end driving policies are commonly developed through a two-step training process: Phase 1 imitation learning and Phase 2 reinforcement/edge-case learning, Fig. \ref{fig:driving_model}. The first stage uses imitation learning to rapidly build a strong baseline by copying competent human driving \cite{paniego2024e2e}. The second stage uses reinforcement learning and edge-case-focused training to systematically improve safety and robustness, especially in rare high-consequence situations that are underrepresented in standard driving data.

\vspace{2mm}

\textit{Phase 1 — Imitation Learning: creating a strong, human-like baseline policy}
In the first stage, a large vehicle fleet collects real-world driving data while humans drive, often described as operating in shadow mode. The model is trained through behavior cloning, learning to predict human control actions such as steering and acceleration from visual input. In addition to raw fleet logs, labeled data and synthetic augmentation are used to expand coverage in variations in weather, lighting, and traffic density. The primary outcome of this stage is a baseline driving policy that performs well in common scenarios.  
Because the objective of imitation learning is to match human actions, the resulting policy tends to behave like a good human driver: it adopts natural lane positioning, comfortable and predictable longitudinal control, socially compliant merging and yielding, and smooth interactions that align with how most drivers expect other vehicles to behave. This is a key reason why E2E systems can feel “natural” to ride with: they internalize the implicit conventions embedded in human driving.
Limitation (the long tail): the same mechanism that makes imitation learning effective under normal conditions also constrains it. The model can struggle with rare or unseen edge cases, simply because the relevant examples are sparse or absent in the dataset, and human responses in unusual situations may be inconsistent.  
In other words, Phase 1 produces a baseline that is broadly competent, but not yet optimized for the safety-critical “long tail.”

\textit{Phase 2 — Reinforcement \& edge-case learning: pushing beyond imitation toward safer behavior}

After establishing a strong baseline, the second stage targets the weaknesses of the policy by explicitly identifying edge cases where the model diverges from the desired driving, including situations where it differs from skilled human behavior or where the baseline is uncertain. Reinforcement learning and other related policy-improvement methods then refine the policy using reward signals that directly encode desirable outcomes: increasing safety margins, smoothness: manage change of speed under changing road conditions, and reduced need for intervention.  
A central advantage of this phase is that it can use simulation and synthetic scenarios to generate additional rare or dangerous situations that would be difficult, slow, or unsafe to capture at scale in the real world. These scenarios allow policy to be tested and improved on events such as abrupt cut-ins, unusual construction configurations, atypical pedestrian behavior, complex right-of-way negotiations, and other low-frequency/high-risk interactions.  This is a way of continuous improvement of the models once a new set of special situations is collected from the fleet of vehicles. Furthermore, this could be an essential mechanism to adapt the system to different cultures around the world. Before deployment, candidate improvements can be validated through fleet replay in shadow mode, where new trajectories are ranked and assessed against safety and comfort criteria.  
The practical outcome is model updates that introduce safer behaviors and improve performance on a new number of rare and critical events. 

\textit{Why Phase 2 can be much safer and generalize better than imitation alone:} Phase 2 differs fundamentally from Phase 1 in what it optimizes:
Imitation learning optimizes similarity to human driving, so it tends to reproduce human-level performance and human-style biases. Reinforcement/edge-case learning optimizes explicit safety and robustness objectives, which can push the policy toward safer-than-human behavior, particularly in scenarios where humans are inconsistent, distracted, or statistically undertrained (rare events).
By repeatedly focusing training on edge cases and rewarding outcomes that reduce risk (collision avoidance, correct yielding, stability, legality, minimal intervention), the system can potentially generalize beyond the patterns most commonly represented in human driving logs. This is the key mechanism by which an E2E policy can evolve from “drives like a good human driver” to “drives in a way that is potentially safer than human drivers,” especially in the long tail.
This second stage also introduces real challenges: reward design, simulation fidelity, and compute constraints can limit how quickly and reliably safety improvements translate to real-world gains. 
Nevertheless, the two-stage structure is powerful: Phase 1 delivers scalable competence and Phase 2 delivers targeted safety and robustness improvements, with particular impact on rare but consequential events.

\section{Robotaxi Deployments: The Road to Level 4 Autonomy}

\textit{A. Waymo}'s commercial robotaxi service launched in Austin, Texas, in March 2025, representing the first Level 4 autonomous ride-hailing operation in that market. Waymo’s vehicles employ a sensor-fusion architecture that combines cameras, LiDAR, and Radar, and have accumulated a substantial safety record in their San Francisco and Phoenix operations. The Austin deployment leveraged Waymo’s established operational design domain (ODD) methodology, beginning with a geofenced service area that was progressively expanded as operational data confirmed the performance of the system in local traffic conditions.
Key metrics from the Austin operation include safety, reliability and passenger satisfaction score that compares favorably with traditional ride-hailing services \cite{charmet2023odd}. Waymo’s approach to safety validation emphasizes extensive simulation testing, structured road trials, and staged geographic expansion governed by quantitative performance thresholds.

\textit{B. Tesla} launched its robotaxi pilot in Austin in June 2025, initially deploying a limited fleet of vehicles operating with a safety driver on the passenger sit. Unlike Waymo’s sensor-fusion approach, Tesla’s robotaxi relies exclusively on cameras, with the FSD E2E architecture providing all perception and planning functions. The absence of LiDAR is a deliberate design choice reflecting Tesla’s claims that camera-only systems trained on sufficient real-world data can match or exceed the performance of sensor-fusion approaches at a fraction of the hardware cost.
The Tesla robotaxi trial attracted significant attention from regulators, industry analysts, and the public, given the higher stakes of fully driverless operation. Tesla used a remote monitoring and intervention capability that allowed operators to observe vehicle behavior and potentially intervene when needed. Early operational data indicated strong performance in the defined service area, and the system exhibited particular strengths in handling complex intersection geometry and mixed traffic environments.

\textit{C. Comparative Analysis and Common Challenges}

Although Waymo and Tesla adopt distinct architectural philosophies, they face a broadly similar set of operational challenges. Both systems must handle degradation of sensor performance in typical weather conditions, navigate construction zones with atypical road geometry, and respond appropriately to unpredictable human and other driver behaviors. Public acceptance remains a key variable, and incidents involving either platform attract disproportionate media attention relative to the far larger number of uneventful miles accumulated.
A critical point of differentiation is scalability. Waymo’s sensor-rich vehicles carry an estimated hardware cost per vehicle that is an order of magnitude higher than that of a comparably capable Tesla, limiting the speed at which Waymo can expand its fleet. Tesla’s vision-only approach, if validated at scale, would represent a substantial cost advantage and could enable robotaxi deployments in markets where the economics of LiDAR-equipped vehicles are prohibitive.

\textit{D. Implications for Scalable Supervised Autonomy}

In general, the Austin deployments align with optimistic expectations. Waymo performed strongly with a sensor-rich vehicle platform, using LiDAR Radar Camera redundancy and tightly controlled ODD \cite{li2020lidar} to provide a reliable Level 4 ride-hailing service. At the same time, Tesla’s FSD demonstrates that a camera-only system running on production hardware can also achieve robust performance within a defined service area, an outcome with important implications for scalability and cost.
Most importantly, Tesla’s results suggest a credible pathway to a widely deployable consumer technology that can perform nearly the full Dynamic Driving Task (DDT) in urban environments while retaining a human supervisor for attention and intervention. In this paper, we refer to this emerging product category as FSD Supervised, often characterized as “L2++”. The commercial and technical traction of supervised E2E autonomy has not gone unnoticed: multiple companies beyond Tesla, most notably NVIDIA and Rivian, have also released strategies and product roadmaps to deploy FSD supervised, which are examined in the following section.

\section{FSD Supervised}

\begin{figure}
    \centering
    \includegraphics[width=\linewidth]{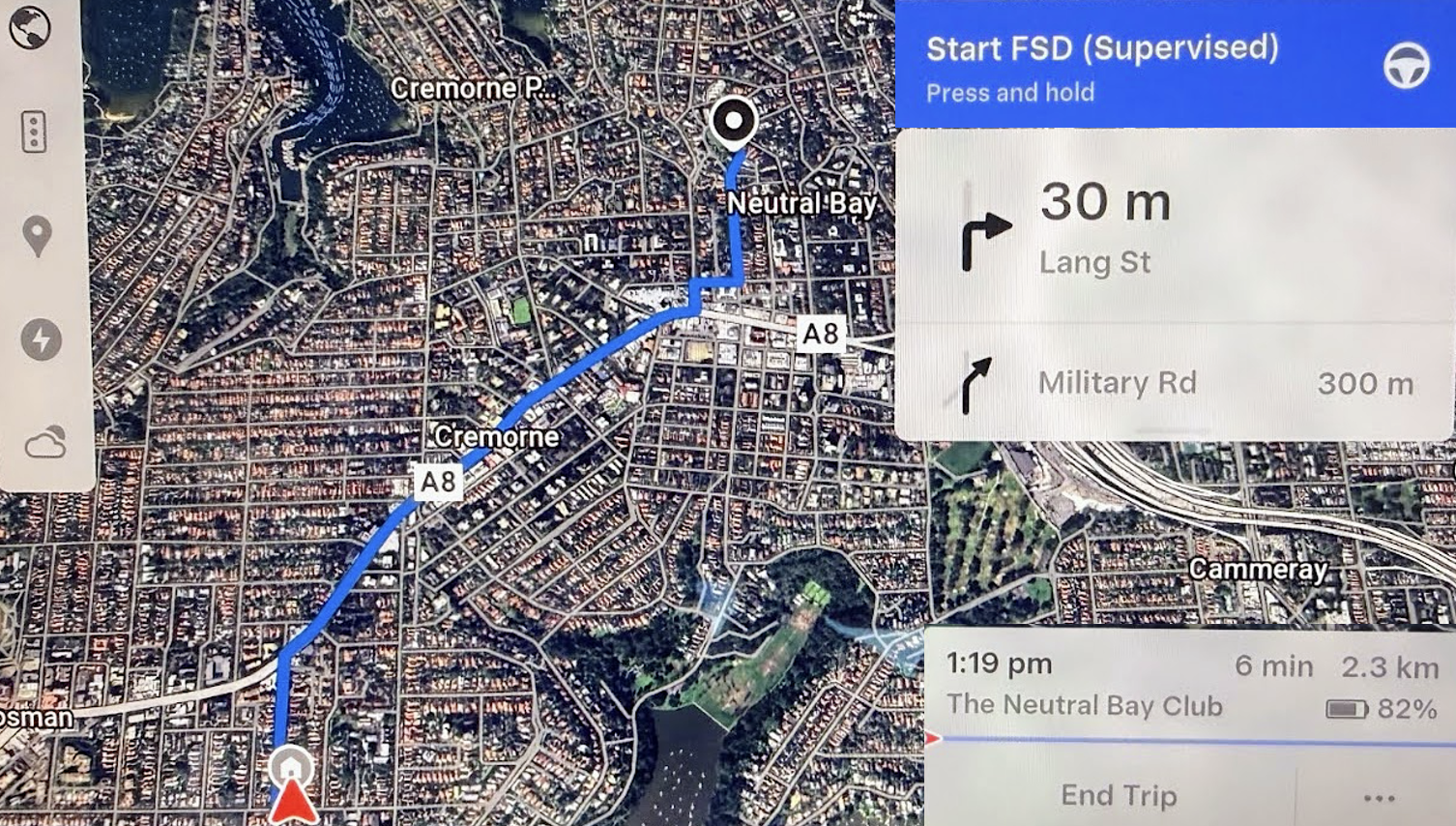}
    \caption{FSD Supervised operation: The driver select the destination and initiate the system; the vehicle will then perform 100\% of the Dynamic Driving Task (DDT) for the entire journey to the destination.}
    \label{fig:Ed_car}
    \vspace{-4mm}
\end{figure}

FSD Supervised is an end-to-end driving mode in which the driver tells the car where to go and the vehicle performs the driving task. In this case, the driver remains responsible as is usually referred to as the supervisor. In practical terms, it is designed so that, once engaged, the car can carry out the 100\% Dynamic Driving Task (DDT) to the selected destination, while the driver continuously monitors the operation and is ready to intervene if necessary, Fig \ref{fig:Ed_car}.
To use it on an urban city trip, the workflow is simple. First, the driver selects a destination in the navigation system, exactly as normal GPS guidance. The driver then starts FSD Supervised; the system takes over the steering, acceleration, braking, lane selection, and intersection handling as it progresses along the route to the destination, Fig \ref{fig:Ed_car}.  

Throughout the drive, the driver’s job is active supervision: keep an eye on the road, understand what the car is doing and be prepared to take control instantly, for example, if the system hesitates, behaves unexpectedly or encounters a complex situation it does not handle well. In other words, the driver is not a passenger; the driver’s role has changed from continuously controlling the vehicle to monitoring and intervening only when needed. 
This becomes a significant paradigm change for driving since in the majority of cases the car will perform 100\% of DDT without any intervention of the driver. The following section describes in detail how the technology works and why it is so natural to the driver.

\subsection{Tesla Full Self-Driving: Architecture and Deployment}
Tesla’s Full Self-Driving (FSD) programme provides one of the clearest real-world examples of the industry’s shift toward end-to-end autonomy. This section summarises the architectural progression from V12 to V14 and links these advances to deployment outcomes, including robotaxi pilots and expansion into right-hand-drive markets.

\subsubsection{FSD V12: The First True End-to-End System}

Tesla’s Full Self-Driving V12, released in 2024, marked a pivotal inflection point not only for Tesla but for the wider autonomy industry, signaling to many developers that end-to-end architectures were becoming a credible path to scalable autonomous driving. For the first time, the entire driving stack from camera inputs to vehicle actuation was replaced by a single neural network, eliminating the thousands of lines of rule-based C++ code that had characterized previous FSD versions. The model was trained end-to-end on driving data collected from the Tesla fleet of millions of vehicles, using human drivers' interventions as a training signal. FSD V12 demonstrated a step-change improvement in natural driving behavior, handling a wider range of scenarios without the abrupt and hesitant maneuvers characteristic of earlier versions.

\subsubsection{FSD V13: Scale and Resolution}
FSD version 13 extended the E2E architecture with higher-resolution video inputs, increasing the effective temporal sampling rate to approximately 36 Hz. This improvement allowed the model to better track fast-moving objects and to exploit finer temporal dynamics in the driving scene. Training data was substantially expanded, including a broader distribution of geographic locations, weather conditions, and traffic patterns. Version 13 also introduced improvements to the model’s handling of unprotected turns, construction zones, and interactions with cyclists and pedestrians, areas that had represented persistent challenges for earlier versions. This model was a significant  step change in performance and was used in the initial robotaxi trials in Austin. This model was also adapted for RHD driving and is currently being deployed in vehicles in Australia and New Zealand.

\subsubsection{FSD V14: Extended Context and Multimodal inputs}
FSD V14 marks the most significant architectural step change so far, with the model scale increasing by roughly a factor of 30 compared to V12. Various Reports claim that  V14 is potentially based on a Mixture of Models (MoM) architecture, in which a learned routing mechanism selects among a set of specialized expert sub-networks depending on the current driving context. This approach allows the model to allocate computational resources efficiently while maintaining specialized competence in diverse scenarios ranging from dense urban intersections to high-speed motorway overtaking maneuvers.
A particularly notable innovation in FSD V14 is the integration of audio processing. The model receives microphone input as an additional sensory modality, enabling it to detect emergency vehicle sirens, construction noise, and other auditory cues that are invisible to cameras but highly informative for driving decisions. The V14 architecture also supports multi-second temporal reasoning, maintaining a compressed latent representation of driving history that informs decisions in scenarios requiring anticipation of future traffic states. Operationally, V14 significantly reduces the frequency of intervention prompts directed at the driver, reflecting greater model confidence and smoother overall performance. This is the model that is currently deployed in USA, Canada, South Korea and being demonstrated in various countries across Europe.

\subsection{Other Similar FSD supervised Technologies}

\subsubsection{Rivian: Fleet Data Flywheel and Large Driving Models}

Rivian’s autonomy roadmap is structured around a fleet learning loop in which real-world operation continuously generates the data required to improve an end-to-end (E2E) driving policy. In the “Rivian Data Flywheel,” customer vehicles operate as a distributed sensing network, producing millions of miles of driving that can be curated for training and evaluation.  A key element of this loop is the Autonomy Data Recorder (ADR), which selectively captures and uploads “important, interesting events,” thereby concentrating bandwidth and labeling effort on the scenarios that matter most for model improvement.  The resulting model is then distilled and rolled out to the fleet via over-the-air (OTA) updates, creating a compound improvement cycle.  

Rivian’s technical narrative combines E2E learning with high-fidelity sensing and centralized computation. Their current Gen 3 autonomy platform is described as using a multi-sensor “trinity,” including 11 Cameras (65 megapixels), a front facing Laser and five Radars, one front imaging and four corner units, aiming to provide robust perception in diverse conditions.  To process the resulting data stream and support increasingly complex models, Rivian emphasizes vertical integration at the compute layer, noting that it has chosen to build its own silicon.  This platform is presented as the foundation for scaling the capacity of the model and reducing dependence on supplier timelines.
At the core of Rivian’s approach is the Large Driving Model (LDM), trained end to-end to map raw sensor input to vehicle trajectory. Importantly, the improvement stage is framed as reinforcement learning aligned not with generic “human values,” but specifically to safe, performant and smooth driving, i.e. optimization towards measurable driving objectives rather than pure imitation. 
Rivian’s public “road to autonomy” is presented as a progression from wide-coverage hands-free driving to point-to-point capability and eventually “personal level 4” operation, potentially in 2026.  
Parallel to the autonomy roadmap, Rivian positions software centralization as strategically important, highlighted by the Rivian–Volkswagen joint venture, which emphasizes movement from scattered ECUs to centralized computing, enabling efficient OTA updates and unification of vehicle functions, including autonomy, under a single software foundation.  Taken together, these elements depict a coherent E2E strategy: a data flywheel to source edge cases at scale, transformer-based trajectory prediction, reinforcement-learning-based refinement toward safety and smoothness, and vertically integrated compute and software distribution to accelerate iteration cycles.

\subsubsection{NVIDIA: World Foundation Models, “Infinite Simulation,” and Dual-Stack Safety with Mercedes}

NVIDIA’s autonomy proposition is built around the concept of physical AI: large-scale models that learn reusable representations of the world and can be applied to prediction, generation, and closed-loop decision-making. NVIDIA developed Cosmos as a “world foundation model” described as the engine that learns the laws of physics.  In this framing, Cosmos supports an infinite simulation loop intended to enable testing and learning at extremely large scale traveling trillions of miles inside a computer by repeatedly generating scenarios and evaluating agent responses in closed loop.  The loop includes a reasoning/action step in which Cosmos “reasons through edge scenarios,” analyzes what could happen next, and decomposes complex events into familiar physical interactions, thereby enabling interactive simulation without leaving the lab.  
Building on this simulation-centric approach, Nvidia developed Alpamayo, also referenced as a thinking autonomous agent, as an end-to-end agent in which the raw sensor input is directly mapped to steering and braking through a single “reasoning model.”  Alpamayo is described as open source and on the order of 10B parameters \cite{NVIDIA2026AlpamayoR1}, suggesting NVIDIA’s intent to catalyze an ecosystem around a reference E2E driving agent rather than limiting innovation to proprietary stacks. It also offers an open dataset with more than 1700 hours of driving. 
A distinctive architectural idea in the NVIDIA approach is a dual-stack safety architecture that runs an E2E policy alongside a more classical deterministic guardrail stack. The “Policy \& Safety Evaluator” monitors confidence in real time and, if confidence drops, switches back to the classical guardrail.  This diversity-and-redundancy design seeks to combine the strong performance of an E2E agent in complex scenes with the predictability and traceability of rule-based safeguards, providing a pragmatic path for incremental deployment under safety constraints.

\subsection{Why the E2E system behaves so natural: Tesla FSD }

Implementations such as Tesla E2E run at high frequencies, in this case 36 Hz, allowing the vehicle to acquire a 360-degree state of the world every 27.7 milliseconds. This rapid sampling and response cycle far exceeds human reaction times, enabling the system to solve immediate problems close to the vehicle with an intuition that feels fluid to the driver.

\begin{figure}
    \centering
    \includegraphics[width=\linewidth]{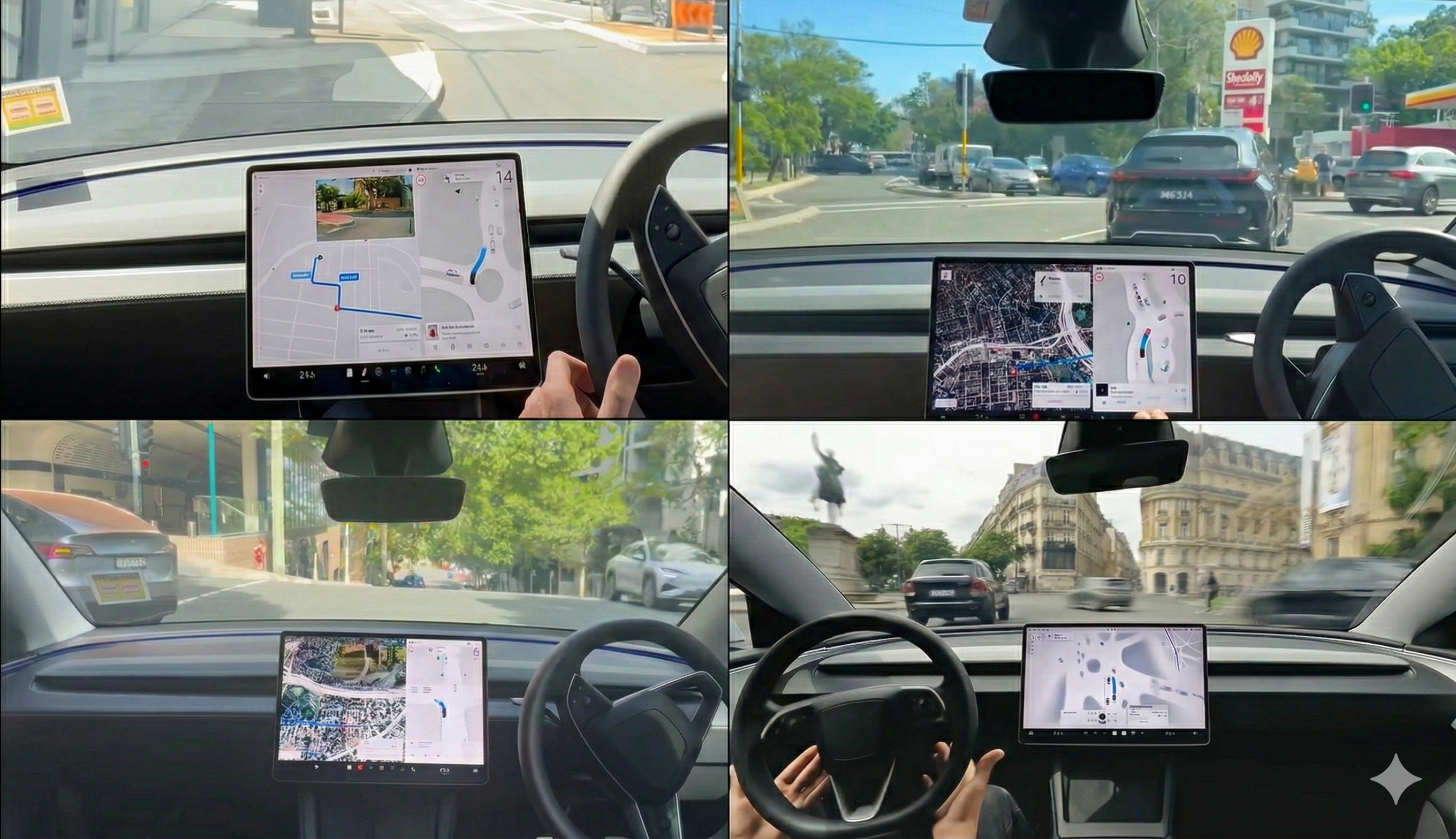}
    \caption{Visualization Interface: The system  provides the driver with clear situational awareness and communicates the vehicle’s immediate intended action and trajectory. (a) Top left: the vehicle will exit the roundabout. (b) Top right: the vehicle will follow the road while turning right. (c) Bottom right: the vehicle will turn right at an intersection. (d) The vehicle will proceed straight, while the interface will display road infrastructure and nearby vehicles to indicate the vehicle’s intended path and surrounding context.}
    \label{fig:tesla_visualisation}
    \vspace{-4mm}
\end{figure}

\begin{figure}
    \centering
    \includegraphics[width=\linewidth]{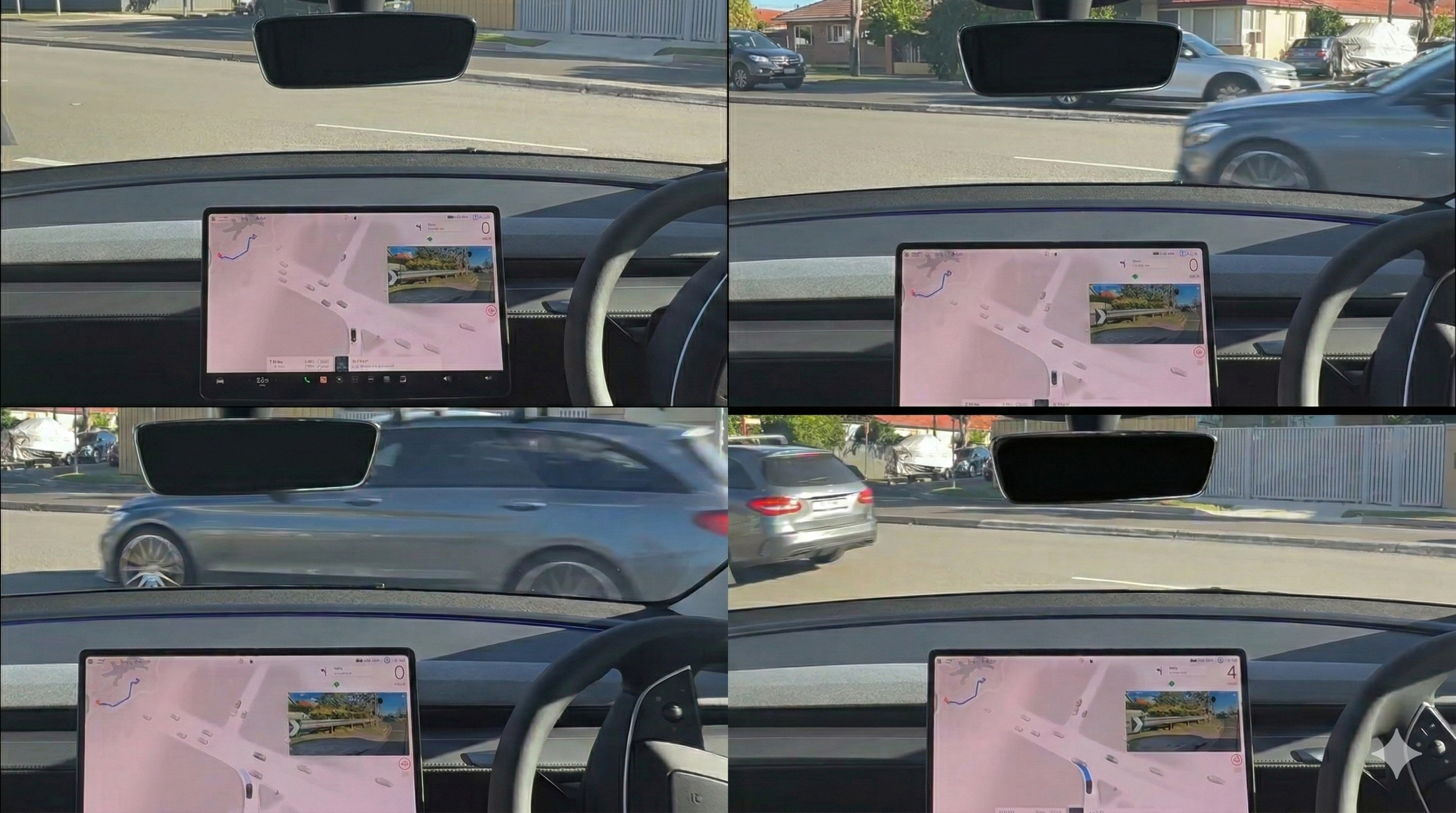}
    \caption{Visual–Haptic–Dynamic interface: (a) Top left: the vehicle is waiting for a gap in traffic to merge. (b) Top right: Haptic—the steering wheel begins to rotate, indicating the vehicle has detected an opening. (c) Top left: Visual—the planned trajectory starts to appear; Haptic—the steering wheel moves; Dynamic—the vehicle accelerates slowly to alert the driver that it is about to move. (d) Visual—the full trajectory is displayed in blue; Dynamic—the vehicle begins the merge.}
    \label{fig:tesla_merge}
    \vspace{-4mm}
\end{figure}

\textit{The Driver Interface for Building Trust: }
A natural feel is not only mechanical; it is also psychological because the driver is constantly forming expectations about what the vehicle will do next. If the system behaves predictably, communicates its intent, and matches common driving conventions, the driver’s mental model aligns with the car’s  \cite{olaverri2020trust}. That alignment reduces uncertainty and anxiety and increases trust during supervision.
The user interface (UI) serves as a bridge, communicating the vehicle's thoughts to the human supervisor. Modern FSD interfaces build trust by providing the following:
 \begin{itemize}
     \item Explainability: High-performing models now predict interpretable outputs, such as 3D occupancy, flow, and traffic semantics.
     \item Predictive Visualization: The user interface communicates the next action of the vehicle before it occurs, reducing passenger anxiety, Fig. \ref{fig:tesla_visualisation} 
     \item Anticipation: In critical maneuvers, the system provides multimodal feedback to the driver and, in some cases, cues to other road users about its intentions. An example is merging onto a main road. Before merging, the system provides progressively richer visual, haptic and dynamic cues, such as slight steering-wheel movement and gentle acceleration, to alert the driver that the vehicle is preparing to move, Fig. \ref{fig:tesla_merge} 
     \item System Reasoning: Advanced agents, such as those pursued by Tesla, NVIDIA, and Rivian, can potentially decompose complex scenarios into plain-language explanations, clarifying why a particular decision (e.g., “gentle braking for a pedestrian”) was made. Significant research is still needed to develop seamless ways of communicating this information to drivers and, where appropriate, to other road users.
 \end{itemize}

In sum, E2E FSD feels natural because it behaves less like a rule-driven machine and more like a skilled driver that executes a continuous stream of small and confident corrections. The high-frequency loop enables immediate responses and smooth control, while the data-driven policy captures the subtle social norms of driving learned from large-scale demonstrations. Reinforcement learning then concentrates on the rare and safety-critical situations where imitation alone is insufficient, producing behaviors that can be more conservative, more consistent, and potentially safer than typical human responses. Finally, the UI closes the trust gap by making intent legible through predictive trajectories, interpretable scene output, and multimodal cues so that the human supervisor can anticipate actions rather than react to surprises. The combination of continuous control, fleet-scale learning, edge-case optimization, and transparent intent communication is what transforms E2E supervised autonomy from technically capable to something that feels intuitively competent to ride with and supervise.

\subsection{Performance boundaries, strategic limitations, and the safety evaluation agenda (Tesla FSD Supervised)}

Tesla’s current E2E implementation can be interpreted through the lens of dual-process decision-making: it exhibits strong System 1 capabilities \cite{Kahneman2011ThinkingFastSlow}, fast, reactive and highly fluent control, while still showing System 2 limitations related to longer horizon planning and strategy. In close range driving, the system performs particularly well in solving immediate problems around the vehicle, supported by high rate sampling (36Hz) and continuous estimation of a 360-degree state of the environment.  This enables rapid reactions after each observation cycle and contributes to smooth lane changes, conservative safety margins, and effective interaction with nearby road users, including pedestrians, bicycles, and motorcycles while maintaining awareness and rule compliance (e.g., speed control, stop signs) and priority behaviors aligned with human values, such as yielding to pedestrians.  
Despite these strengths, the system can still improve in System 2 functions that require anticipatory planning over longer horizons and complex route intent. Reported limitations include route planning that is logically correct but sometimes difficult to execute under heavy traffic, late preparation for multi-lane positioning (e.g., needing to enter a main avenue shortly before multiple lane changes to turn), and inconsistent early lane selection for right- and left-hand turns particularly in RHD countries, where right-hand turns present distinctive challenges.  Additional behaviors noted for improvement include speed sign interpretation and speed selection, often conservative and sometimes failing to account for variable speed time zones, which can cause the vehicle to obey a lower posted limit when contextual timing rules should apply.

\subsection{FSD Supervised: Deployment and Geographic Expansion}
Tesla’s FSD Supervised product makes E2E autonomous driving available to consumers while retaining the requirement of driver attention and readiness to intervene. By early 2026, FSD Supervised had accumulated tens of billion miles of operation across North America, Canada, South Korea and had been deployed in right-hand-drive (RHD) markets, most notably Australia/New Zealand. The RHD deployment represents a significant engineering achievement, as it requires the model to generalize to mirror-image road geometry, different road markings, and different traffic conventions. 
The deployment also has important implications for original equipment manufacturers (OEMs) beyond Tesla. The availability of independent FSD providers and licensable LDM-based solutions is expected to accelerate industry-wide adoption of E2E architectures through 2026 and 2027. Rivian and Nvidia-Mercedes CLA will introduce FSD supervised in 2026 in US, Europe, and potentially Asia. Furthermore, This technology could also be incorporated in Volkswagen vehicles as part of the joint venture with Rivian.
In summary, 2026 is likely to mark the start of large-scale consumer deployment of supervised E2E driving across multiple OEMs, moving the technology from a single-vendor offering toward a broader industry capability. This expansion will potentially not be “one-size-fits-all”: each rollout must be adapted to country-specific road geometry, signage and markings, traffic rules, driving culture, mapping conventions, and regulatory requirements, including differences between the markets of left- and right-hand drive. As the deployment scales, the main limiting factor will increasingly shift from model capability alone to safe operational integration, notably user training, clear supervision responsibilities, and consistent human machine interaction design that teaches drivers how to monitor, when to intervene and how to interpret the intent of the vehicle. These requirements imply that successful adoption in 2026–2027 will depend as much on localization and education programs as on continued improvements in end-to-end model performance.

\subsection{Safety in FSD (Supervised): metrics, methodology, and evidence}

Unlike Level-4 robotaxi services, FSD (Supervised) safety depends on (i) the ability of the end-to-end policy to execute the Dynamic Driving Task (DDT) and (ii) the human supervisor’s ability to monitor and intervene when required \cite{olaverri2020tor}. A defensible safety case, therefore, treats driver monitoring, interface design, and intervention dynamics as part of the system.

\textit{Interpreting safety through baselines and outcome severity:} A rigorous evaluation of safety must specify the baseline (e.g., average human driving) and separate outcomes by severity, minor vs major collisions.  This is essential because supervised E2E systems can reduce certain types of crash while leaving others largely unchanged, and because early rollouts can be biased toward easier roads, better weather, or more attentive users.
Obtaining the baseline is not a simple exercise. The limited set of recent safety disclosures from industry generally benchmark performance against average human driving, rather than against matched vehicle fleets or comparable driver populations \cite{izquierdo2024pedestrian, hussien2025rag, morales2021gaze}. This average includes bad drivers, drunk drivers, etc. At the same time although we want a system to perform better than a good driver, we recognize that the demographics of drivers include drivers with very different skills and different ages. Furthermore, it is also important to consider the type of technology included in a car when evaluating the baseline. Nevertheless, the fact that we start to see updated safety performance information from companies becomes a very important first step.

\textit{Where the system fails vs where it excels:} For supervised E2E, the most useful safety evidence is not only collisions but also precursors of an accident: near misses, harsh braking, unsafe gap acceptance, and late or abrupt maneuvers that force other road users to react. In parallel, the assessment should explicitly document: (i) where the system is prone to issues, (ii) where it outperforms a normal driver, and (iii) how these regions could potentially change with software updates.  

\textit{Interpreting Tesla’s reported collision-rate snapshot:} Tesla reports miles driven before a collision comparisons that indicate higher miles-per-collision when FSD (Supervised) is engaged than for broader benchmarks. The chart shows approximately 5.1 million miles per major collision with FSD (Supervised) engaged versus about 0.7 million for a U.S. benchmark, and approximately 1.5 million miles per minor collision versus about 0.23 million for the same benchmark.  According to their conclusions, the system is approximately 7 times safer than an average driver.

\begin{figure}
    \centering
    \includegraphics[width=\linewidth]{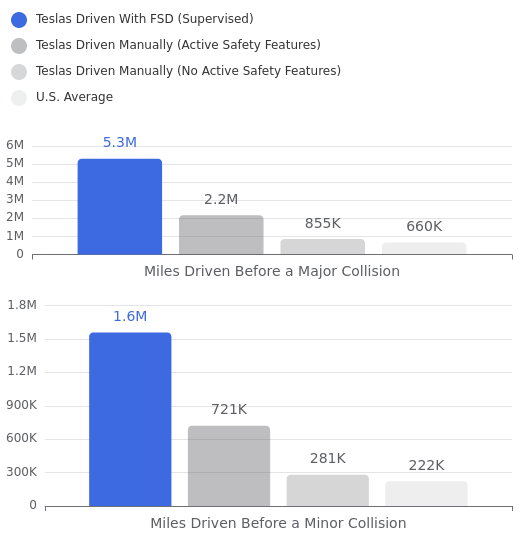}
    \caption{Miles driven before major/ minor collisions (non-highway, worldwide), showing higher distances between collisions for Tesla vehicles using FSD (Supervised) compared to manual driving and the U.S. average. \cite{templeton2025tesla}}
    \label{fig:tesla_crashes}
    \vspace{-4mm}
\end{figure}

These figures should be interpreted as indicative rather than definitive safety evidence, unless they are accompanied by transparent normalization for exposure and selection effects (e.g., who uses the feature, where and when it is used, operating conditions, collision definitions, and reporting methodology. In addition, the comparison to an average driver considers many older cars that do not have any safety technology. A case could be made that a more appropriate comparison could be made with Teslas driven manually with actual safety features. In this case, the result is still safer but 1.7 times instead of 7. Nevertheless, the fact that we start to see updated information on safety performance from companies becomes a very important step \cite{TeslaSafetyReport}

\textit{Who benefits most in terms of segmentation and policy relevance:} The safety impact should be evaluated by user segment (e.g., older adults, young drivers, commuters, families, people with disabilities), since the baseline risk, exposure, and supervision capacity vary substantially between populations.  Segmentation is also important for public policy because the net societal benefit of supervised E2E autonomy may be highest where it reduces high-risk exposure without introducing supervision risk.

In summary, a credible safety case for FSD (Supervised) must evaluate the combined human–automation system, using clearly defined baselines, severity-stratified outcomes, and leading indicators such as interventions and near-miss proxies. Tesla’s published fleet statistics provide useful early signals and help frame the discussion, but robust conclusions require careful normalization of exposure and fair comparisons between vehicle capabilities and driver populations. As supervised E2E deployment expands, safety assessment will increasingly depend on transparent reporting, continuous monitoring throughout the long tail, and segmentation to identify where technology delivers the greatest net benefit.

\section{Beyond Vehicles: End-to-End Intelligence in Robotics}
Architectural and training advances that have accelerated end-to-end (E2E) autonomy in vehicles large-scale imitation learning, reinforcement learning for edge cases, high-frequency closed-loop control, and fleet-scale data pipelines are increasingly being transferred to robotics. The underlying problem can be considered similarly: mapping high-dimensional sensory inputs to safe and stable actions under uncertainty in complex open-world environments. This section outlines two representative trajectories toward scalable embodied intelligence: Tesla’s transfer of autonomy-style E2E learning to humanoids and NVIDIA’s foundation-model approach with simulation at scale, as a transferable technology for robot fleets.

\subsection{Tesla Optimus}
The same architectural innovations driving progress in AVs are being applied to humanoid robotics. Tesla’s Optimus program develops bipedal robots trained using E2E imitation learning from human demonstrations and reinforcement learning in simulation. The robot uses a camera-based perception system that shares architectural elements with FSD, and its motor control policies are trained end-to-end on manipulation and locomotion tasks. By 2025, Optimus had been deployed in Tesla’s Fremont factory performing quality inspection and light assembly tasks, and Tesla had announced plans to manufacture the robot on a scale for external customers.

\subsection{NVIDIA Project GR00T}
NVIDIA’s Project GR00T (Generalist Robot 00 Technology) is a foundation model initiative for humanoid robots. GR00T aims to provide a generalized sensorimotor policy that can be fine-tuned to specific robot platforms and tasks, analogous to the role played by large language models as a foundation for downstream natural language applications. The model is trained on a combination of human motion capture data, simulation episodes generated by NVIDIA’s Isaac Sim platform, and real-world robot demonstrations. GR00T’s architecture is explicitly designed to leverage NVIDIA’s Cosmos physical world simulator, which generates photorealistic, physics-accurate training data at scales that would be prohibitively expensive to collect in the real world.
Overall, these efforts suggest a convergence toward data-driven end-to-end sensorimotor policies that can be improved iteratively through a combination of real-world experience and simulation. As these systems mature, key research challenges will focus on long-term robustness, safety validation, and effective human–robot interaction during supervision and deployment.

\section{Conclusion}

This paper has argued that autonomous driving is entering an era in which end-to-end (E2E) learning and Large Driving Models (LDMs) are displacing traditional modular pipelines as the dominant engineering and commercial strategy. The limitations of sense–perceive–plan–control stacks, particularly brittleness in the long tail, high integration burden, and reliance on expensive sensors and map maintenance, have created strong incentives for unified learned policies trained at fleet scale. In contrast, modern E2E systems leverage two complementary training stages: imitation learning to achieve good human driver competency and reinforcement learning to concentrate improvement on edge cases, safety objectives, and robust generalization. This shift is no longer theoretical, it is reflected in product deployments and roadmaps from Tesla, FSD (V12–V14), Rivian’s LDM program and NVIDIA’s physical-AI ecosystem, and it is reinforced by the emerging alignment between autonomy providers and OEM partners.
A key conclusion is that the near-term inflection point of the industry is the supervised E2E autonomy, here referred to as FSD Supervised or L2++. The robotaxi pilots discussed in this paper illustrate a credible pathway: sensor-heavy Level 4 systems can deliver strong performance inside tightly defined operational design domains, while camera-only approaches running on production vehicles can also achieve robust operation in constrained deployments. The latter is strategically significant because it implies a cost structure compatible with mass-market vehicles. If validated and scaled, supervised E2E systems could execute most of the Dynamic Driving Task in urban environments, shifting the human role from continuous operator to safety supervisor, a change with implications comparable to other major transitions in transportation technology.  
However, wide deployment depends on more than model capability. The primary bottlenecks are increasingly operational and social: (i) localization to country-specific road rules, signage, infrastructure, and driving culture, including left vs  right-hand drive conventions, (ii) consistent human–machine interface design that makes intent legible, and (iii) user education and supervision protocols that reduce misuse and clarify responsibility. In parallel, safety assessment must mature from headline collision-rate comparisons toward a transparent framework that combines severity-stratified outcomes with leading indicators: interventions, near misses, conflict metrics, exposure normalization, and segmentation across user groups and operating conditions. Manufacturer fleet statistics are valuable early signals, but robust conclusions require careful methodology and comparability.
Finally, the paper highlights that the same architectural advances that power E2E driving, high-frequency closed-loop control, fleet-scale data pipelines, reinforcement learning for rare events, and simulation at scale are now accelerating robotics. As autonomy and robotics converge toward general-purpose physical AI, the research agenda expands: scalable safety validation, sim-to-real transfer, interpretable intent communication, and implementations of continuous OTA updates will be as important as model accuracy. The evidence reviewed here supports a forward-looking conclusion: 2026–2027 is likely to be the beginning of a broad commercialization of supervised E2E autonomy, with success determined by a combination of technical robustness, safety governance, localization and human adaptation.


\bibliographystyle{ieeetr}
\bibliography{bibtex/bib/IEEEexample}

\end{document}